\documentclass[10pt,twocolumn,letterpaper]{article}

\usepackage[pagenumbers]{wacv} % To force page numbers, e.g. for an arXiv version

\usepackage{graphicx}
\usepackage{amsmath}
\usepackage{amssymb}
\usepackage{booktabs}
\usepackage{ dsfont }

\usepackage[pagebackref,breaklinks,colorlinks]{hyperref}

\usepackage[capitalize]{cleveref}
\crefname{section}{Sec.}{Secs.}
\Crefname{section}{Section}{Sections}
\Crefname{table}{Table}{Tables}
\crefname{table}{Tab.}{Tabs.}

\def\wacvPaperID{30} % *** Enter the WACV Paper ID here
\def\confName{WACV - GeoCV Workshop}
\def\confYear{2025}

\begin{document}

\title{CrossModalityDiffusion: Multi-Modal Novel View Synthesis with Unified Intermediate Representation}

\author{
    Alex Berian, Daniel Brignac, JhihYang Wu, Natnael Daba, Abhijit Mahalanobis\\
    University of Arizona: ECE Dept.\\
    Tucson, AZ\\
{\tt\small [berian, dbrignac, jhihyangwu, ndaba, amahalan]@arizona.edu}
}

\maketitle

%%%%%%%%% ABSTRACT
\begin{abstract}
Geospatial imaging leverages data from diverse sensing modalities—such as EO, SAR, and LiDAR, ranging from ground-level drones to satellite views. These heterogeneous inputs offer significant opportunities for scene understanding but present challenges in interpreting geometry accurately, particularly in the absence of precise ground truth data. To address this, we propose CrossModalityDiffusion \footnote{Code: \href{https://github.com/alexberian/CrossModalityDiffusion/}{https://github.com/alexberian/CrossModalityDiffusion/}}, a modular framework designed to generate images across different modalities and viewpoints without prior knowledge of scene geometry. CrossModalityDiffusion employs modality-specific encoders that take multiple input images and produce geometry-aware feature volumes that encode scene structure relative to their input camera positions. \textbf{The space where the feature volumes are placed acts as a common ground for unifying input modalities. }These feature volumes are overlapped and rendered into "feature images" from novel perspectives using volumetric rendering techniques. The rendered feature images are used as conditioning inputs for a modality-specific diffusion model, enabling the synthesis of novel images for the desired output modality. In this paper, we show that \textbf{jointly training different modules ensures consistent geometric understanding across all modalities} within the framework. We validate CrossModalityDiffusion's capabilities on the synthetic ShapeNet cars dataset, demonstrating its effectiveness in generating accurate and consistent novel views across multiple imaging modalities and perspectives.
\end{abstract}

\section{Introduction}
\label{sec:intro}

\begin{figure}[t]
  \centering
   \includegraphics[trim=180 100 180 100, width=0.47\textwidth, clip]{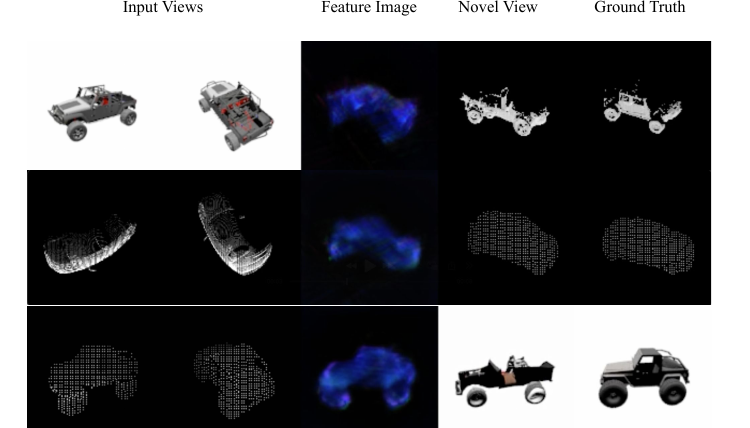}% left bottom right top

   \caption{\textbf{CrossModalityDiffusion input/output examples.} Three examples of CrossModalityDiffusion with two input images, and an output from a different modality. From the top down: EO to SAR, LiDAR(RA) to LiDAR(P), LiDAR(P) to EO. The corresponding intermediate feature image is shown in the middle, and the ground truth target image is shown on the right.}
   \label{fig:preview}
\end{figure}

Geospatial imaging has become increasingly abundant, with datasets spanning a wide range of image sensing modalities such as electro-optical (EO) \cite{eurosat}, synthetic aperture radar (SAR) \cite{terrasarx}, and Light detection and ranging (LiDAR) \cite{geolidar}. Each sensing modality uniquely captures critical features of the Earth's landscape, making them indispensable for comprehensive scene analysis. However, understanding the underlying geometry of scenes from sparse images from various modalities remains a challenging problem.

Neural Radiance Fields (NeRF) \cite{nerf} revolutionized novel view synthesis (NVS) by learning scene geometry exclusively from images, enabling high-quality view generation. However, NeRF's reliance on dense input images and inability to extrapolate beyond observed viewpoints limits its utility in many scenarios. In contrast, GeNVS \cite{genvs} explicitly models scene geometry, allowing it to infer structure from sparse inputs. GeNVS uses an encoder to produce geometry-aware feature volumes from each input image, which are aligned and rendered via volume rendering to create a feature image for a novel view. This feature image is then processed by a diffusion model \cite{ddpm, edm} to generate the final output.

While NeRF and GeNVS have shown remarkable results for NVS with EO images, they work on a single imaging modality. Extending NVS to multiple imaging modalities introduces the problem of multi-modal novel view synthesis (MMNVS), where input and output images may belong to different imaging modalities. Existing approaches like GeNVS face significant limitations in MMNVS due to their dataset-specific nature, requiring retraining for each new modality. Additionally, their tightly coupled architecture restricts generalization.

In this work, we present CrossModalityDiffusion, a modular framework designed for MMNVS. We validate CrossModalityDiffusion on the ShapeNet cars \cite{shapenet} dataset, which we render in EO, perspective LiDAR (LiDAR(P)), range-angle LiDAR (LiDAR(RA)) and SAR modalities. Our results demonstrate that CrossModalityDiffusion effectively synthesizes accurate and consistent novel views across these diverse imaging modalities, showcasing its capability for generalized MMNVS.

In Figure \ref{fig:preview} we observe CrossModalityDiffusion operating on EO to SAR, LiDAR(RA) to LiDAR(P), and LiDAR(P) to EO. We see from the feature images in Figure \ref{fig:preview} that \textbf{no matter the input modality, CrossModalityDiffusion produces geometrically consistent feature volumes.}

CrossModalityDiffusion decouples the three primary components of the GeNVS architecture — (1) the encoder, (2) feature image rendering, and (3) the denoising diffusion model — allowing independent modality-specific modules for input and output modalities. Namely, we use specific encoders and denoisers for each image modality. The feature volumes produced by the encoders are modality-agnostic, allowing for modality fusion. By jointly training the modules within the framework, we ensure the encoder modules learn consistent and transferable geometric representations for the different imaging modalities.

\section{Related Work}
\label{sec:related_work}

NeRF \cite{nerf} revolutionized novel view synthesis by implicitly training a multi-layer perception (MLP) to map 3D coordinates and view directions to color and density for volume rendering. While NeRF achieved state-of-the-art (SOTA) quality in NVS, it requires a dense set of input images to render photo-realistic views. Additionally, NeRF is scene-specific and cannot generalize to unseen targets. Many works incrementally improved NeRF in image quality \cite{mipnerf,zipnerf,refnerf,kplanes}, training time \cite{initnerf,kplanes}, and extended its capabilities \cite{initnerf,kplanes,neus,sonarneus}.  We particularly focus on few-shot NVS, the problem of generating novel views of a scene when given one or a small number of source images. PixelNeRF \cite{pixelnerf} addressed this problem by training an image encoder with a NeRF module for scene-agnostic performance. GeNVS \cite{genvs} utilizes a PixelNeRF-like pipeline to produce geometry-aware priors for a powerful diffusion model \cite{edm} for few-shot NVS.

Diffusion models are now the standard in high-quality image generation \cite{originaldiffusion,ddpm,iddpm,scorebased,edm} by iteratively denoising the raw image space. Latent diffusion models (LDMs) \cite{edm2,ldm} instead iteratively denoise a latent space representation of data, efficiently generating high-quality images. DreamFusion \cite{dreamfusion} presented score-distillation-sampling (SDS), which showed that a web-scale text-to-image latent diffusion model \cite{ldm} implicitly learned 3D information. SDS uses a fixed diffusion model as a critic to train a NeRF. SparseFusion \cite{sparsefusion} adopts SDS for few-shot NVS by conditioning the LDM with a view-aligned feature grid. Zero-1-to-3 \cite{zero123,zero123plus} extends the concept in SparseFusion to re-train the LDM with camera pose embeddings to produce better SDS scenes. DreamGaussian \cite{dreamgaussian} uses the concept of SDS for Gaussian splatting \cite{gaussiansplatting} scenes instead of a NeRF.

More recently, various works \cite{reconfusion,cat3d} show that when a diffusion model outputs better novel views, those images can be used to directly train a NeRF to get better results than SDS. Multi-view diffusion models, first presented in MVDream \cite{mvdream}, utilize cross attention \cite{transformer,vit} over multiple views of a scene to train a more 3D-aware diffusion model. CAT3D \cite{cat3d} uses a multi-view diffusion model trained at a web-scale to directly produce images for training a NeRF on any scene.

Planar scene representation methods \cite{3planegan, kplanes,lrm,triposr,sf3d}, similar to PixelNeRF \cite{pixelnerf}, project render points onto planes to interpolate feature vectors as input to a NeRF model. EG3D \cite{3planegan} trains a Style-GAN \cite{stylegan}, to predict 3 planes to represent scenes and fixed NeRF to render the scenes. Large reconstruction models \cite{lrm,triposr,sf3d} use a similar concept to EG3D but utilize a large visual transformer \cite{vit} instead of a GAN.

\section{Problem Statement}
\label{sec:prob}
NVS is the problem of inferring a target image $x_t \in \mathds{R}^{(128,128,3)}$ from $S$ source images $\{x_{si}\}_{i=1}^S$\footnote{For brevity, we denote $\{x_{si}\}_{i=1}^S = \{x_{s1}, x_{s2}, x_{s3}, \dots, x_{sS}\}$}. The target images and source images have associated camera pose matrices $P_t \in \mathds{R}^{(4,4)}$ and $\{P_{si}\}_{i=1}^S$ respectively. We seek to create a NVS model $N_\theta$ with parameters $\theta$ that predicts novel views

\begin{equation}
    \hat{x}_t = \ N_\theta(P_t,\{x_{si}\}_{i=1}^S,\{P_{si}\}_{i=1}^S)
    .
    \label{eq:nvs}
\end{equation}

MMNVS extends the problem of NVS to where the images may come from $M$ different modalities. Each source and target image has an associated modality, $(x_{si},m_{si})$ and $(x_t,m_t)$ where $m\in\{1,2,...,M\}$. In this paper, we assume modality information is known to the model, so NVS expression in (\ref{eq:nvs}) can be rewritten for MMNVS as 
\begin{equation}
    \hat{x}_t = \ N_\theta(P_t,m_t,\{x_{si}\}_{i=1}^S,\{P_{si}\}_{i=1}^S,\{m_{si}\}_{i=1}^S)
    .
    \label{eq:mdnvs}
\end{equation}

\section{Background: GeNVS}
\label{sec:bkgd_genvs}

GeNVS \cite{genvs} encodes each source image $x_{si}$ image into feature volumes $W_i \in \mathds{R}^{128,128,64,16}$ (point clouds of 16-dimensional features) using a modified DeepLabV3+ \cite{deeplab,deeplabv3p,segmentationgithub} segmentation model. The volumes are oriented within the camera field-of-view (frustrum) of their source camera poses $P_{si}$ and a dataset-specific $z_{near}$ to $z_{far}$.

A 16-channel feature image $F \in \mathds{R}^{64,64,16}$ of the feature volumes is rendered via volume rendering from a target camera pose $P_t$. Stratified sampling is used to select points along each ray $\mathbf{r}$. A point $r \in \mathbf{r}$ is trilinearly interpolated in each feature volume to get a latent vector $W_i(r) \in \mathds{R}^{16}$. The latent vectors for each feature volume are averaged together, then passed through a MLP $f$ to get a 16-channel color $c$ and density $\sigma$.
\begin{equation}
    c(r),\sigma(r) = f\Bigg(\sum_{i=1}^S W_i(r)\Bigg)
    .
    \label{eq:genvs_avg}
\end{equation}
Once the color and density of every point along a ray $\mathbf{r}$ are predicted, volume rendering is used to predict the 16-channel pixel value $C(\mathbf{r})$. The ray rendering formula follows the NeRF \cite{nerf} equations
\begin{equation}
    C(\mathbf{r}) = \sum_{i=1}^{N}T_i(1-e^{-\sigma_i\delta_i})\mathbf{c_i}
    ,
\end{equation}
where, $\sigma_i$ and $\mathbf{c}_i$ are the MLP output of the $i^{th}$ sample along the ray from the camera, where

\begin{equation}
    T_i =  e^{-\sum_{j=1}^{i-1}\sigma_j\delta_j}
    ,
\end{equation}
and $\delta_i=t_{i+1}-t_i$ is the distance between the $i^{th}$ and $(i+1)^{th}$ samples.

The 16-channel feature image is used as conditioning input to a denoising diffusion \cite{edm} U-Net $U$ by concatenating it with 3-channel noise (or noisy target image in training). The denoising U-Net $U$ outputs 3 channels. When sampling new images with multiple denoising steps, the feature image is skip-concatenated to the U-Net's $U$ input. The final output target image $\hat{x}_t$ is the output from the last denoising step. 

\section{Method}
\label{sec:method}

\subsection{Multi-Modal Dataset}
To better understand how we adapt GeNVS for MMNVS, we must first examine the dataset used to train the framework.

For the model to generalize to unseen images from unseen scenes, we need to first collect images from a variety of scenes. Then from each scene, we need to take a lot of images of it from different viewpoints. From each viewpoint, we need to use different sensors to capture the scene (EO, LiDAR, SAR, etc).

\begin{figure}
  \centering
   \includegraphics[trim=0 192 479 0, width=0.47\textwidth, clip]{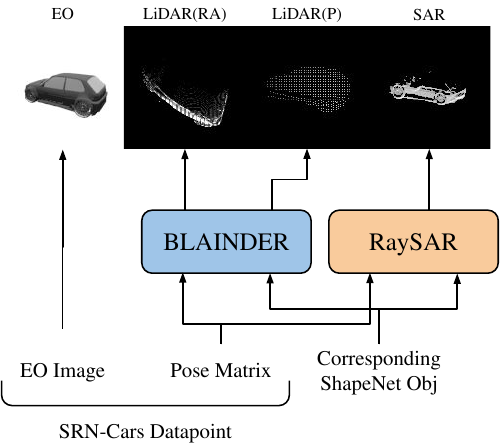}% left bottom right top

   \caption{\textbf{Dataset Generation.} We begin with the SRN-Cars dataset for EO images and corresponding camera pose matrices. We then use the pose matrix and ShapeNet object file for generating LiDAR(RA) and LiDAR(P) images with BLAINDER, and SAR images with RaySAR}
   \label{fig:dataset}
\end{figure}

In our experiments, we use the ShapeNet Cars dataset. We get the EO and corresponding poses from the SRN-Cars \cite{srn} dataset. We then use BLAINDER \cite{blainder} to generate LiDAR(RA)/LiDAR(P) images and RaySAR \cite{raysar} to generate SAR images. This is shown in Figure \ref{fig:dataset}.

\subsection{Architecture}
\begin{figure}
  \centering
   \includegraphics[trim=0 70 374 0, width=0.47\textwidth, clip]{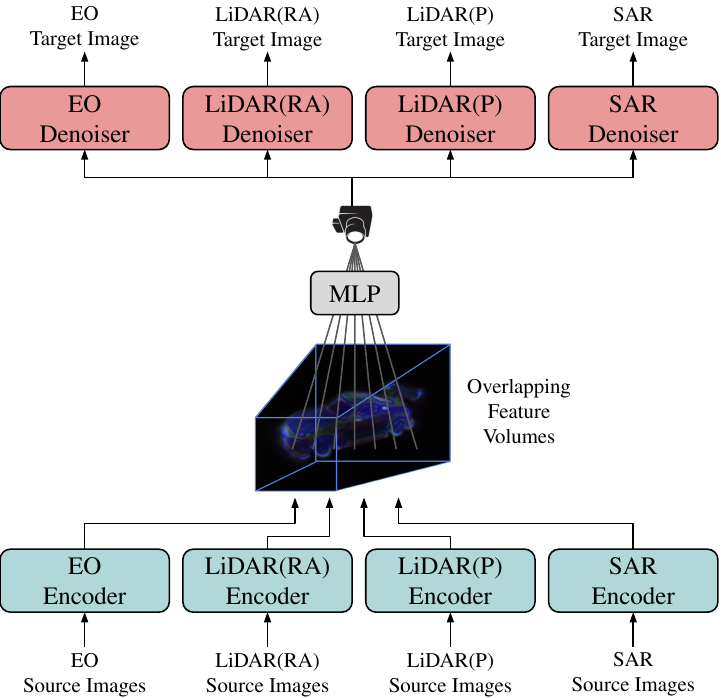}% left bottom right top

   \caption{Architecture of CrossModalityDiffusion.}
   \label{fig:architecture}
\end{figure}

GeNVS has three main components. The encoder, feature image generation, and the denoising diffusion model. We begin building the model by training an unmodified GeNVS architecture from scratch exclusively on EO images. This allows for faster learning when training on other modalities. Once the EO-only GeNVS model is pre-trained, we initialize a new encoder and denoiser module for each input and output modality from the pre-trained EO encoder. This architecture is shown in Figure \ref{fig:architecture}.

GeNVS is composed of three major modules: the encoder, the MLP, and the denoiser. By allocating an encoder and denoiser module for each modality, we obtain a modular framework of adapters for MMNVS.

To unify the intermediate representation from the encoders, we only use one MLP to process the overlapping feature volumes created by the different modality encoders. Then according to the target modality, we select the appropriate denoiser for conditional diffusion.

\subsection{Joint Training}
\label{sec:joint}
To allow our framework to input multiple images from different modalities at the same time then output a novel view of the scene in any modality, we jointly train the encoder and denoiser modules for each modality at the same time. 

During training, we randomly select between one and three input images from random views and modalities, as well as one random target view and modality. This way, the encoders are incentivized to generate the same feature field despite the different source image modalities. By jointly training the different modules with random input and target modalities, we can create a unified, modality agonistic, implicitly learned intermediate representation of the scene.

Any encoder can encode an image of its modality into the intermediate representation and any denoiser can decode any feature image of a novel view into an image of its modality.

% \subsection{Range-angle Images}
% Not all sensor modalities produce perspective-projection alike images. We challenge our framework to also work on range-angle images with little modification.

% Although our framework works with range-angle images without any modification, we found it beneficial to use a different approach when transforming the feature field into a feature image.

% Instead of using the volume rendering equation to compose the feature image, we explicitly calculate the range and angle of each $H \times W \times D$ sampled point. We then plot the queried feature vectors onto 
% the feature image by accumulating their values on the nearest 4 pixels and weighting using bilinear interpolation weights and density $\tau$ from the MLP.

% TODO: add equations here

% While volume rendering techniques work great for perspective-projection-like images, range-angle images work better with this method as the shape of the feature image matches geometrically with the novel view.

\section{Experiments}
\label{sec:experiments}

We follow the distribution of train/validation/test as in the SRN-Cars \cite{srn} dataset. EO data is directly taken from the SRN-Cars dataset, and we render the same viewpoints in other modalities. We use RaySAR \cite{raysar} and BLAINDER \cite{blainder} for generating SAR and LiDAR images respectively. At present, we use two modes of LiDAR images in the experiments: range-angle (RA), and perspective (P) images. For LiDAR and SAR images, we repeat the single-channel image across three channels for compatibility purposes.

All experiments are conducted on fixed modules trained with the joint training process described in section \ref{sec:joint}. Although the evaluation metrics used are primarily used for EO images, they also function as a useful metric for MMNVS.

\subsection{Single Modality In, Single Modality Out}
Results when all input images to CrossModalityDiffusion are of the same modality are shown in Table \ref{tab:shapenet_benchmark}. There is only one output image at a time from the framework, so it is trivial to say all output images are the same modality.

We observe from Table \ref{tab:shapenet_benchmark} that the easiest of these tasks is EO in EO out (as shown by the 19.66 PSNR). This is because volume rendering captures color information, but not depth. The toughest tasks are any other modality in EO out. This is because the input images from non-EO modalities contain no color information. Images of a black and white car with otherwise consistent geometry would yield a large pixel difference. 

\begin{table}[h!]
\centering
\resizebox{0.47\textwidth}{!}{
\setlength{\tabcolsep}{4pt} % Reduce column spacing
\begin{tabular}{@{}l@{\hspace{0.5em}}c@{\hspace{0.5em}}c@{\hspace{0.5em}}c@{\hspace{0.5em}}c@{\hspace{0.5em}}c@{\hspace{0.5em}}c@{\hspace{0.5em}}}
\hline
In Modality & Out Modality & \begin{tabular}[c]{@{}c@{}}FID↓\end{tabular} & \begin{tabular}[c]{@{}c@{}}LPIPS↓\end{tabular} & \begin{tabular}[c]{@{}c@{}}DISTS↓\end{tabular} & \begin{tabular}[c]{@{}c@{}}PSNR↑\end{tabular} & \begin{tabular}[c]{@{}c@{}}SSIM↑\end{tabular} \\
\hline
EO & EO & 19.69 & 0.13 & 0.18 & 19.66 & 0.87 \\
EO & LiDAR (RA) & 29.20 & 0.19 & 0.21 & 17.23 & 0.77 \\
EO & LiDAR (P) & 17.00 & 0.13 & 0.17 & 17.20 & 0.77 \\
EO & SAR & 34.24 & 0.20 & 0.20 & 16.49 & 0.85 \\
LiDAR (RA) & EO & 52.07 & 0.26 & 0.27 & 14.52 & 0.80 \\
LiDAR (RA) & LiDAR (RA) & 28.90 & 0.18 & 0.20 & 17.66 & 0.78 \\
LiDAR (RA) & LiDAR (P) & 16.14 & 0.12 & 0.16 & 17.46 & 0.78 \\
LiDAR (RA) & SAR & 33.42 & 0.19 & 0.20 & 16.64 & 0.85 \\
LiDAR (P) & EO & 44.40 & 0.25 & 0.26 & 14.76 & 0.80 \\
LiDAR (P) & LiDAR (RA) & 24.97 & 0.15 & 0.19 & 19.06 & 0.81 \\
LiDAR (P) & LiDAR (P) & 13.97 & 0.09 & 0.15 & 17.51 & 0.79 \\
LiDAR (P) & SAR & 29.56 & 0.17 & 0.19 & 17.46 & 0.86 \\
SAR & EO & 50.15 & 0.26 & 0.26 & 14.48 & 0.80 \\
SAR & LiDAR (RA) & 27.35 & 0.18 & 0.20 & 17.52 & 0.78 \\
SAR & LiDAR (P) & 16.05 & 0.11 & 0.16 & 17.41 & 0.78 \\
SAR & SAR & 31.81 & 0.18 & 0.19 & 17.17 & 0.86 \\
\hline
\end{tabular}
}
\caption{\textbf{Single modality in, single modality out.} We evaluate CrossModalityDiffusion on ShapeNet Cars dataset. To quantify generated image quality, we calculate the Fréchet Inception Distance \cite{fid} (FID), Learned Perceptual Image Patch Similarity \cite{lpips} (LPIPS), Deep Image Structure and Texture Similarity \cite{dists} (DISTS), Peak Signal-to-Noise Ratio (PSNR), and Structural Similarity Index Measure (SSIM). ↑ or ↓ indicate better performance from higher or lower values respectively.}
\label{tab:shapenet_benchmark}
\end{table}

\subsection{Range-angle Images}
Not all sensor modalities produce perspective-projection alike images. We challenge our framework to also work on range-angle images with no modification. We find the denoiser trained on range-angle images still performs well despite no geometry-aware rendering of feature images.

% https://github.com/JhihYangWu/OurGeNVS/blob/041a63055b87291e2e04cf14628a876559e78cc8/genvs/models/pixel_nerf_net.py#L112
However, we also found it beneficial to render the feature image as a range-angle image. We explicitly calculate the range and angle of all the sampled points and plot it on an empty image. Since the calculated range and angle may lie between pixels, we use bilinear interpolation weights to scale the feature based on how close it is to each of the four nearest pixels.
% TODO

\subsection{More Images, Better Results}

Modality fusion is the process of combining data from different source modalities into the model to provide more reliable, accurate, and useful information. This is a common and important problem in autonomous systems that have multiple sensing modalities such as EO, LiDAR, radar, ultrasound (and so forth) from different locations and orientations. By jointly training the various modality-specific modules of CrossModalityDiffusion, we can fuse information of different modalities from multiple viewpoints into a single, more informative, feature volume representation of the scene.

In Table \ref{tab:data_fusion}, we evaluate our framework on ShapeNet Cars by randomly selecting  $S$ input views from random modalities and comparing the average output image quality (also from random modalities) against when only one of those input images is used. The main conclusion from this experiment is that when more images are used as input, the output image quality improves. This is because CrossModalityDiffusion is fusing its geometric understanding of the scene from each input image. We also observe that as the number of input views increases (lower on Table \ref{tab:data_fusion}) results improve. Note that the upper number in each row of Table \ref{tab:data_fusion} does not change much, because the CrossModalityDiffusion is always using one input image.

\begin{table}[h!]
\centering
\resizebox{0.47\textwidth}{!}{
\setlength{\tabcolsep}{4pt} % Reduce column spacing
\begin{tabular}{@{}l@{\hspace{0.5em}}c@{\hspace{0.5em}}c@{\hspace{0.5em}}c@{\hspace{0.5em}}c@{\hspace{0.5em}}c@{}}
\hline
Method & \begin{tabular}[c]{@{}c@{}}FID↓\end{tabular} & \begin{tabular}[c]{@{}c@{}}LPIPS↓\end{tabular} & \begin{tabular}[c]{@{}c@{}}DISTS↓\end{tabular} & \begin{tabular}[c]{@{}c@{}}PSNR↑\end{tabular} & \begin{tabular}[c]{@{}c@{}}SSIM↑\end{tabular} \\
\hline
$S$ = 2 input views\\
Separate & 29.99 & 0.20 & 0.21 & 16.40 & 0.79 \\
Fused & \textbf{27.47(-8.4\%)} & \textbf{0.18(-10.9\%)} & \textbf{0.20(-5.3\%)} & \textbf{16.83(+2.7\%)} & \textbf{0.81(+1.6\%)} \\
\hline
$S$ = 3 input views\\
Separate & 28.92 & 0.20 & 0.22 & 16.43 & 0.79 \\
Fused & \textbf{26.03(-10.0\%)} & \textbf{0.17(-14.7\%)} & \textbf{0.20(-7.8\%)} & \textbf{17.15(+4.3\%)} & \textbf{0.81(+2.2\%)} \\
\hline
$S$ = 4 input views\\
Separate & 29.36 & 0.20 & 0.21 & 16.38 & 0.80 \\
Fused & \textbf{25.04(-14.7\%)} & \textbf{0.16(-16.8\%)} & \textbf{0.19(-9.0\%)} & \textbf{17.22(+5.1\%)} & \textbf{0.82(+2.6\%)} \\
\hline
$S$ = 5 input views\\
Separate & 29.23 & 0.20 & 0.21 & 16.52 & 0.80 \\
Fused & \textbf{25.38(-13.2\%)} & \textbf{0.16(-18.3\%)} & \textbf{0.19(-9.8\%)} & \textbf{17.48(+5.8\%)} & \textbf{0.82(+2.9\%)} \\
\hline
\end{tabular}
}
\caption{\textbf{Many Vs. One Input image.} In each row the non-bold number for each metric indicates when just one random source image from a random modality is used as input. Note that this number does not change across rows, we simply observe variance. The second number result in bold is when $S$ (indicated by the top left number in the row) random images from random modalities are used as input. For convenience, the number in parenthesis is the difference between the one-input and the $S$-input results.}
\label{tab:data_fusion}
\end{table}

\subsection{Benefits of Having a Variety of Sensors}

\begin{table}[h!]
\centering
\resizebox{0.47\textwidth}{!}{
\setlength{\tabcolsep}{4pt} % Reduce column spacing
\begin{tabular}{@{}l@{\hspace{0.5em}}c@{\hspace{0.5em}}c@{\hspace{0.5em}}c@{\hspace{0.5em}}c@{\hspace{0.5em}}c@{}}
\hline
Modalities & \begin{tabular}[c]{@{}c@{}}FID↓\end{tabular} & \begin{tabular}[c]{@{}c@{}}LPIPS↓\end{tabular} & \begin{tabular}[c]{@{}c@{}}DISTS↓\end{tabular} & \begin{tabular}[c]{@{}c@{}}PSNR↑\end{tabular} & \begin{tabular}[c]{@{}c@{}}SSIM↑\end{tabular} \\
\hline
LiDAR (RA) & 32.343 & 0.218 & 0.224 & 15.911 & 0.784 \\
SAR & 32.204 & 0.211 & 0.220 & 16.042 & 0.788 \\
LiDAR (P) & 41.537 & 0.183 & 0.206 & 16.690 & 0.805 \\
LiDAR (RA) \& SAR & 30.697 & 0.197 & 0.213 & 16.332 & 0.796 \\
LiDAR (RA) \& LiDAR (P) & 28.536 & 0.180 & 0.204 & 16.749 & 0.807 \\
SAR \& LiDAR (P) & \textbf{28.387} & 0.181 & 0.204 & 16.730 & 0.807 \\
LiDAR (RA) \& SAR \& LiDAR (P) & 28.542 & \textbf{0.179} & \textbf{0.203} & \textbf{16.770} & \textbf{0.808} \\
\hline
\end{tabular}
}
\caption{\textbf{Different Sensing Modalities at Same Viewpoint.} Results when the input images are all of the same viewpoint with the specified modalities. Random modality outputs are used for this experiment. We observe that more input sensing modalities yields better performance. This is specifically seen in the bottom row of the table.}
\label{tab:same_viewpoint}
\end{table}

Integrating the strengths of various sensing modalities into a unified, more informative representation is highly advantageous in geospatial imaging. As shown in Table \ref{tab:same_viewpoint}, leveraging our framework with multiple sensing modalities captured from the same viewpoint enhances performance in the MMNVS task.

In this experiment, we evaluate every subset of \{LiDAR (RA), SAR, LiDAR (P)\} sensing modalities. For each test scene, we randomly select one view as input and three views to predict. The subset of sensing modality images serves as input to CrossModalityDiffusion, which generates novel views for evaluation. We observe the best results in the bottom row of Table \ref{tab:same_viewpoint}. As expected, our framework achieves the best performance when all sensing modalities are utilized, even when taken from the same viewpoint. This result underscores the complementary nature of different sensing modalities; each captures unique details of the scene that, when combined, contribute to a richer and more accurate representation.

Another observation we make from Table \ref{tab:same_viewpoint} is that as the number input views increases, the model is using more parameters. In the bottom row of Table \ref{tab:same_viewpoint} more input modalities are used, hence more of the encoder modules as shown in Figure \ref{fig:architecture} are used. The same cannot be done with just one input modality, because the model would use the same weights multiple times.

\subsection{Qualitative Discussion}

\begin{figure*}[p]
    \centering
    \includegraphics[trim=120 25 120 25, clip, width=\textwidth]{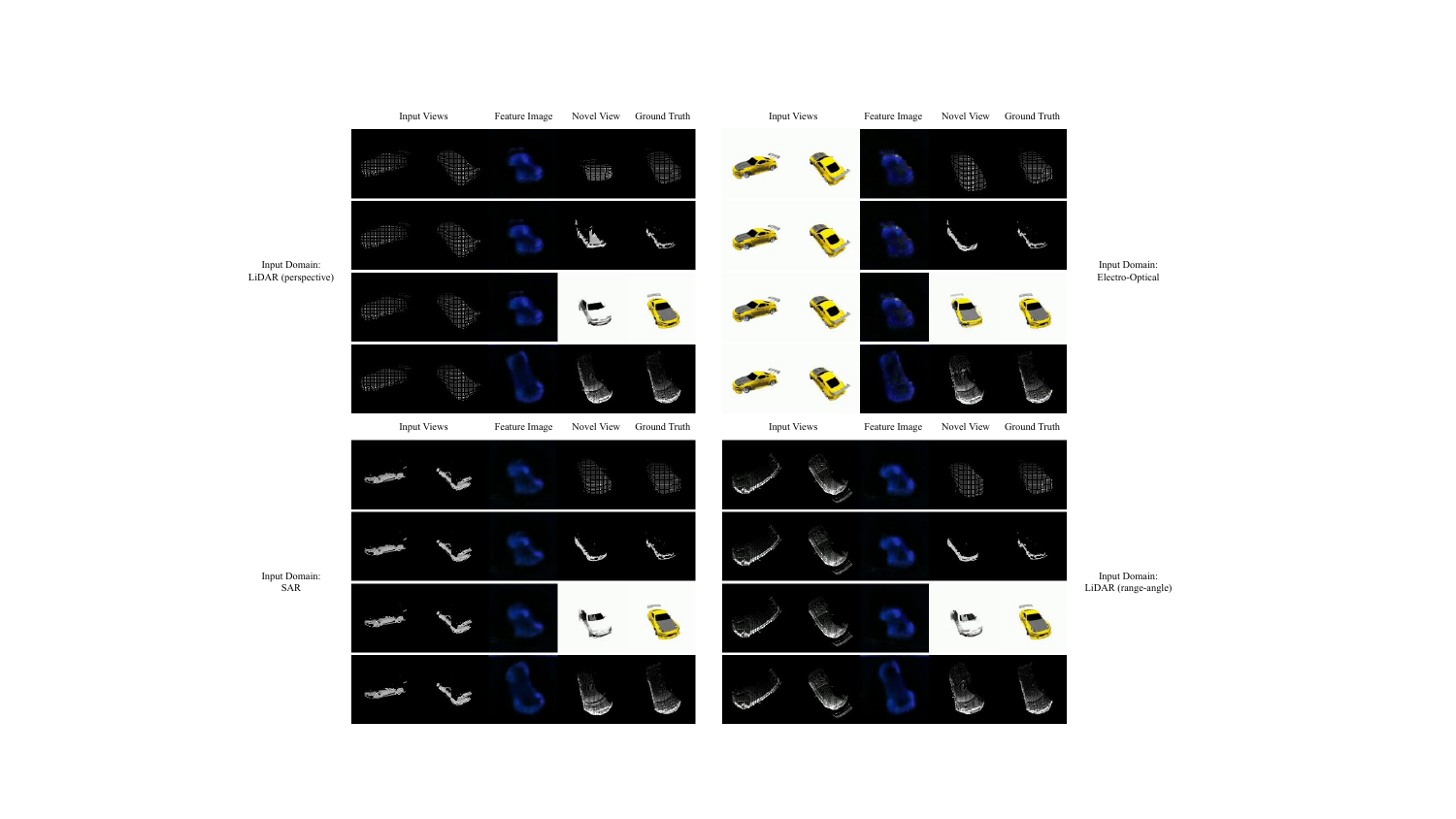}  % left bottom right top
    \caption{Demonstration of any modality to any modality through unified intermediate representation.}
    \label{fig:images}
\end{figure*}

In Figure \ref{fig:images} we show images from all input/output modalities with two inputs from the same modality. We confirm our hypothesis that EO out from other modalities in is difficult because there is no color information. We see in the EO out examples in Figure \ref{fig:images} that the denoiser outputs a white car when the car is actually yellow. EO to EO does not have this problem.

\section{Conclusion}
\label{sec:conclusion}

We introduced CrossModalityDiffusion, a modular framework for few-shot MMNVS using a unified intermediate representation. Our experiments demonstrate that CrossModalityDiffusion can effectively transform images across modalities and integrate information from diverse sensing modalities to enrich scene understanding. By jointly training multiple GeNVS models, we ensure that the learned intermediate representation remains consistent and modality-agnostic. Furthermore, we show that our framework can be adapted to handle non-perspective-projection-based images with minimal modifications.

While CrossModalityDiffusion performs well in MMNVS, its computational demands for training and inference are significant, primarily due to its reliance on a diffusion-based backbone. Despite this limitation, we hope CrossModalityDiffusion inspires new directions in MMNVS and data fusion across sensing modalities. We believe the modality-agnostic intermediate representation has the potential for other downstream tasks beyond MMNVS. Additionally, CrossModalityDiffusion can address scenarios where data availability is imbalanced across sensing modalities, serving as a bridge to overcome such challenges.

%%%%%%%%% REFERENCES
\clearpage
{\small
\bibliographystyle{ieee_fullname}
\bibliography{egbib}
}

\end{document}